# A Novel Approach to the Diagnosis of Heart Disease using Machine Learning and Deep Neural Networks

**Sahithi Ankireddy**[*]

[*] James B. Conant High School, Hoffman Estates, IL, USA
Email: sahithia14@gmail.com

*Abstract*- Heart disease is the leading cause of death worldwide. Currently, 33% of cases are misdiagnosed, and approximately half of myocardial infarctions occur in people who are not predicted to be at risk. The use of Artificial Intelligence could reduce the chance of error, leading to possible earlier diagnoses, which could be the difference between life and death for some. The objective of this project was to develop an application for assisted heart disease diagnosis using Machine Learning (ML) and Deep Neural Network (DNN) algorithms. The dataset was provided from the Cleveland Clinic Foundation, and the models were built based on various optimization and hyper parametrization techniques including a Grid Search algorithm. The application, running on Flask, and utilizing Bootstrap was developed using the DNN, as it performed higher than the Random Forest ML model with a total accuracy rate of 92%.

*Keywords*- deep neural networks, heart disease, Keras, machine learning, sci-kit learn

## I. INTRODUCTION

Heart Disease is the leading cause of death for both men and women. About 610,000 people die of heart disease in the United States every year which rounds out to 1 in every 4 deaths [1]. A conclusive and early diagnosis of heart disease could be the difference between life and death for some. However, one in 3 heart disease cases are misdiagnosed causing patients to miss out on early treatment options [2]. With the number of heart disease patients expected to rise in the near future, it is vital to find a solution. The use of Artificial Intelligence (AI) and more specifically Machine Learning (ML) techniques and Deep Neural Networks (DNN) can mitigate the possibility of human error while increasing prediction accuracy rates. The expected outcome of the use of these data analytic techniques is a higher accuracy prediction rate of at least 75% or more. Additionally, it is expected for the DNN to have a higher accuracy rate because of its ability to back propagate and as theory states. All in all, the model, whether it is the ML algorithm or DNN, with the best accuracy will be used to create an application that reads required data inputs for patients to determine an accurate heart disease diagnosis. This tool can possibly be a great contribution to the cardiology field as it can be used by medical care professionals to assist them in more accurate diagnoses.

## II. BACKGROUND INFORMATION

With the rise in big data, machine learning has become a key method and technique for solving problems in various areas such as computational biology and finance, computer vision, aerospace and manufacturing. ML is a general data analysis technique that uses computational models and methods to "learn" information directly from data without using a preset formula or rule-based programming. ML algorithms teach and train computers to recognize patterns and correlations between data. The algorithms are able to improve their performance as the number of data sets increases. A subset of machine learning includes deep learning or the training of a deep neural network. "DNNs are a set of algorithms, modeled loosely after the human brain, that are designed to recognize patterns. They interpret sensory data through a kind of machine perception, labeling or clustering raw input [3]. DNN layers are composed of nodes. Nodes are locations for where a computation happens, similar to a neuron in a human brain. DNNs take in the desired number of inputs and starts off by randomly assigning a weight to each input, that either amplify or lessen the value of the input. Then, the products of all the connections to the node one is computing the activation of are added. This sum is then put through a specific linear algebra function, such as a sigmoid function, as chosen by the programmer. All in all, each input goes through a series of matrix multiplication computations through a linear algebra function to get to its final output, in a process known as forward propagation. The series of matrix computations are called hidden layers (See Figure 1). The number of hidden layers varies based on the programmer and the problem. If there is more than one hidden layer, the results given from the previous hidden layer serve as the input values for the next set of calculations [4]. Another important part in deep neural networks are loss function. Loss functions are used to measure the disparity between predicted values and the actual labels. It is a non-negative value, and the lower the loss function value is, the more robust a model [5].

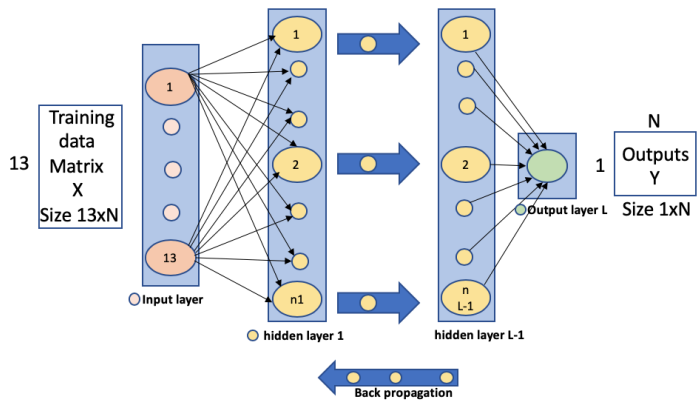

Figure 1. A depiction of the functionality of a deep neural network.

ML and DNNs can be classified into two main groups (supervised and unsupervised learning) based on the way they learn. Supervised learning depends on labeled data sets. Labeled data sets rely on individuals with domain knowledge to manually label the data. Consequently, the DNN is able to learn the patterns and correlations between the data and the labels. Additionally, in this type of learning, DNN's are able to use back propagation and alter the weights of each input so that a computed response can match the target value or output. Supervised learning uses two main techniques: classification and regression. Classification algorithms produce discrete responses or binary data. There are only two outputs, usually in the form of "yes or no" or "1 or 0". Regression algorithms predict continuous responses. For example, this includes weather data (changes in temperature) or changes in height of a child as he/she grows older.

Unsupervised learning is when there are no labels within the data set so the DNN must learn by identifying special features and components. Unsupervised learning utilizes one main technique: clustering. Clustering algorithms organize unlabeled data into similar groups called clusters. Data inside a cluster is similar to each other but are distinctly different from the other clusters.

A common category of machine learning algorithms is ensemble learning. Ensemble methods combine several decision trees classifiers to produce better predictive performance than a single decision tree classifier. The main concept behind the ensemble model is where a group of weak learners come together to form a strong learner, thus, increasing the accuracy of the model.

### III. MATERIALS AND METHODS

#### A. Materials

Data preprocessing, training, and testing was all completed in the Python programming language via an interactive python Jupyter notebook. Furthermore, Anaconda, a package manager system, had to be downloaded to set up the environment.

A publicly available supervised data set provided by the Cleveland Clinic Foundation was used for the ML model and to train the DNN. This data set contains 75 total attributes of patient medical information for 303 patients. 14 attributes out of the 75 were chosen. See Figure 2 for the chosen attributes, and its information (if applicable). These attributes have been selected by other researchers and healthcare professionals because they are known to be the best determining factors of heart disease.

| | Attributes | Information |
|---|---|---|
| 1 | age | age in years |
| 2 | sex | 1=male , 0= female |
| 3 | cp (chest pain type) | 1=typical angina, 2=atypical angina 3= non-anginal pain, 4= asymptomatic |
| 4 | trestbps (resting blood pressure) | in mm Hg on admission to hospital |
| 5 | chol (cholesterol) | serum cholesterol in mg/dl |
| 6 | fbs (fasting blood sugar) | 1= true, 0=false |
| 7 | restecg (resting electrocardiographic results) | 0= normal , 1= ST-T wave abnormality 2= showing probable or definite left ventricular hypertrophy |
| 8 | thalach (maximum heart rate achieved) | |
| 9 | exang (exercise included angina) | 1= yes , 0=no |
| 10 | oldpeak (ST depression induced by exercise relative to rest) | |
| 11 | slope (slope of peak exercise ST segment) | 1= upsloping, 2=flat, 3= downsloping |
| 12 | ca (number of major vessels colored by flouroscopy) | |
| 13 | thal (thallium heart scan results) | 3=normal, 6=fixed defect, 7= reversable defect |
| 14 | num (patient diagnosis of heart disease and predicted attribute) | 0=unlikely to obtaih heart disease 1=likely to obtain heart disease |

Figure 2. A chart of all the features in dataset.

In general, for both algorithms, the machine learning portion and the deep neural network, certain libraries were imported. Pandas was used to read and analyze the data. Seaborn and matplotlib were imported for data visualization techniques. For the machine learning model, the scikit learn library was utilized, whereas the deep neural network was engineered though Keras, a neural network framework.

#### B. Data Preprocessing

The dataset was first loaded into the memory. Next, the original Cleveland data set was split into 2 sets: train and test data set, and each new file was assigned to a variable: *trainData* and *testData*. Additionally, since the prediction of heart disease is a classification problem with two classes, a new function was created to replace any numbers greater than 1 in the last column of the data set (diagnosis column) with 1. Next, the train and test datasets are further separated into the 13 attributes and the predicted diagnosis, to create a total of 4 new sets of data: *Xtrain, Ytrain, Xtest, Ytrain.* The "x" data sets represent the 13 values that play a part in determining if a patient has heart disease. The "y" data sets represent the final outcome, for those patients (1 or 0).

#### C. Project Overview

Before specific details, Figure 3 below illustrates the entirety of this project in a simple graphical way.

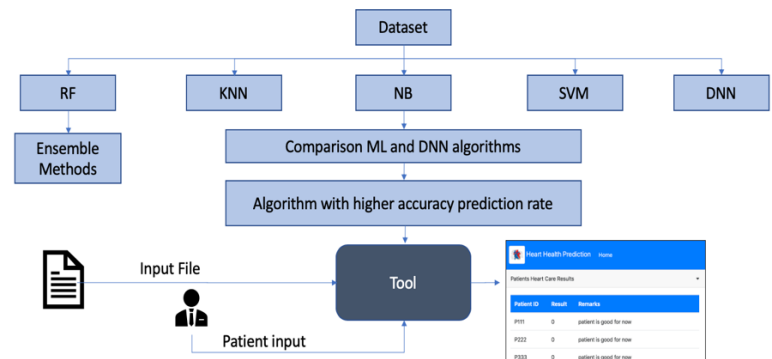

Figure 3. Flow chart of project.

### D. Machine Learning Model Generation and Testing

First, to determine which algorithm to choose, a series of the most common ML algorithms, including K-Nearest Neighbors (KNN), Support Vector Machine (SVM), Random Forest Classifier (RF), Naive-Bayes (NB) were defined. Then these models were fitted with the train data. The model will be trained or fitted on the loaded data by calling the fit () function of the model.

After the models have been created, they were scored based on a series of methods to determine which ML algorithm would work best for the data set. The first scoring method utilized was simply accuracy score. Each model predicted outputs with the Xtest set as inputs and the predictions were compared with the actual results for the dataset, Ytest.

The Receiver Operating Characteristic, or ROC, curve was graphed as well in which it is a plot of the true positive rate versus the false positive rate for the predictions of a model for multiple thresholds between 0.0 and 1.0. The integrated area under the ROC curve, called AUC or ROC AUC, is a measure of the skill of the model across all evaluated thresholds. From the ROC graph a ROC AUC score was also determined. These techniques were not able to be applied to the SVM due to the functionality of the algorithm.

Finally, the Matthews Correlation Coefficient score was calculated for these 4 models, to represent another way to score and evaluate the models. After these 3 scoring methods, it was determined that Random Forest had the best results.

Random Forest is an ensemble method algorithm. In hopes of truly finding the best possible ML algorithm for this certain data set, all the ensemble method classification algorithms from scikit learn were also evaluated. The same method of splitting the data and defining/fitting the models was applied to the AdaBoost Classifier, Bagging Classifier, Extra Trees Classifier, and Gradient Boosting Classifier. The accuracy score was determined, and Random Forest performed the highest compared to the accuracy score of the other ensemble method models, proving the conclusion that Random Forest Model was best ML algorithm for this data set.

Optimization techniques and more specifically a Grid Search Algorithm was applied to the Random Forest Model to find the best combination of hyperparameters.

### E. Deep Neural Network Generation and Testing

Moving on to the DNN, hyperparameters such as number of nodes, activation function, learning rate, number of hidden layers, epochs, and batch sizes were tuned until the best combination was found. Plus, the data features were scaled on a range of 0 to 1 in order for the neural network to better learn from the data. Like with the ML algorithms, the DNN model will be complied and trained/ fitted on the loaded data by calling the fit () function of the model.

Next, the performance of the network was evaluated via the test data. The DNN will take in the 13 values (*Xtest*) and predict the outcomes for the patients. This prediction will be compared with the actual outcomes in the test data (*Ytest*), through loss functions in order to find the accuracy of the model.

### F. Web Application

Once all of the models have been evaluated and finalized, the application was developed through Flask. The DNN was more successful than the machine learning algorithm, so thus it will be the one used in the application.

The application was created using the Flask REST API, where the user interface was developed using Bootstrap. First, the model was exported and saved into a pickle file. Logins and file upload features were added. After the user uploads a csv file with patient data, it's put into a location defined in the program. Once saved to the correct location, the data is passed to the DNN for heart disease diagnosis. Additionally, the patient ID column was initially removed as it's not needed for the diagnosis through the model. At the end, it was later appended for displaying the results.

### IV. RESULTS AND DISCUSSION

In total for DNN and ML algorithm, 297 records were used each with 14 total attributes. 13 out of the 14 features played a part in determining the last feature or the heart disease diagnosis. The original Cleveland data set was split into train data with 236 records and test data with 61 records.

In determining which ML algorithm to use, accuracy scores were produced for each of the common machine learning algorithms. Figure 4 shows the accuracy scores of the K-Nearest Neighbors, Support Vector Machine, Random Forest Classifier, and Naive-Bayes Algorithms, and proves how the Random Forest had the highest accuracy score.

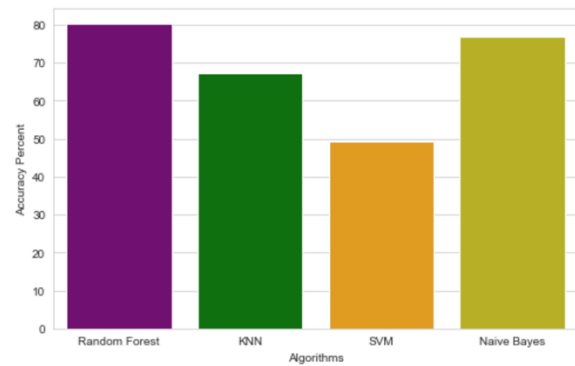

Figure 4. Accuracy Scores of 4 Machine Learning Algorithms.

ROC graphs were also established for RF, KNN, and NB with the AUC score below as shown respectively in Figure 5, 6, 7.

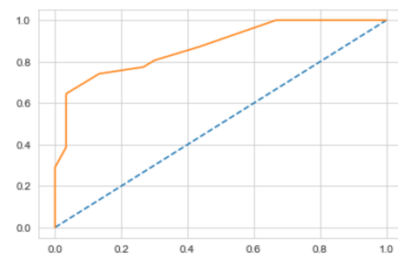

0.8715053763440861

Figure 5. ROC AUC for RF.

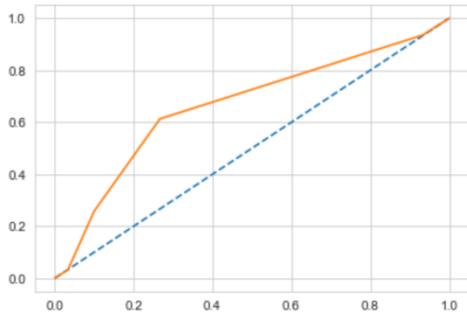

Figure 6. ROC AUC for KNN.

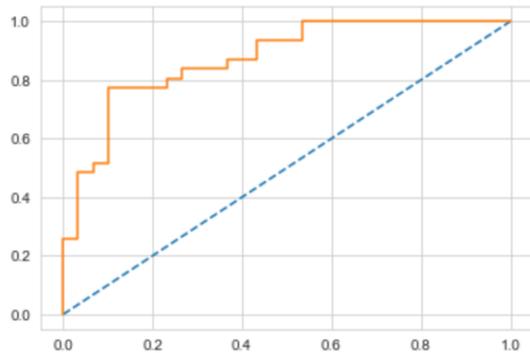

Figure 7. ROC AUC for NB.

A numerical way to view the correlation between false positives, true positives, false negatives, and true negatives is through the Matthews Correlation Coefficient (MCC). MCC is a correlation coefficient between target and predictions. It generally varies between -1 and +1. -1 when there is perfect disagreement between actuals and prediction, 1 when there is a perfect agreement between actuals and predictions. See Figure 8 below.

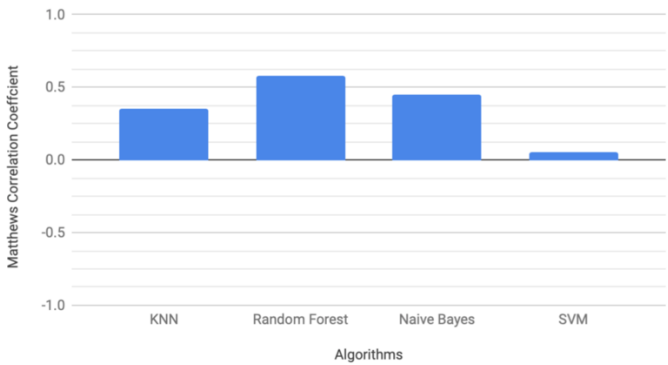

Figure 8. Graphical chart of the MCCs for the ML algorithms.

For ROC graphs, models that have skill have a curve above the diagonal line that bows towards the top left corner. A more solidified way to measure this can be through the AUC score. An AUC score of 0.5 suggests no skill, e.g. a curve along the diagonal, whereas an AUC of 1.0 suggests perfect skill. The Naive Bayes and Random Forest Models had the highest AUC scores and very similar, with Naive Bayes being the highest, and technically the better model in this test. However, looking at the MCC score and accuracy score, Random Forest was the winner. Despite slightly falling short in the AUC score, to Naive Bayes, Random Forest proved to be the better overall model as seen by its much higher performance when looking at accuracy score and Matthews Correlation Coefficient. Thus, it can be deemed that it was the best algorithm out of the 4 tested.

To further prove this conclusion, all the ensemble classification methods in scikit learn were also tested and evaluated via accuracy score. As seen below in Figure 9, Random Forest did once again prove to perform the best, and therefore was chosen as the ML algorithm for this project.

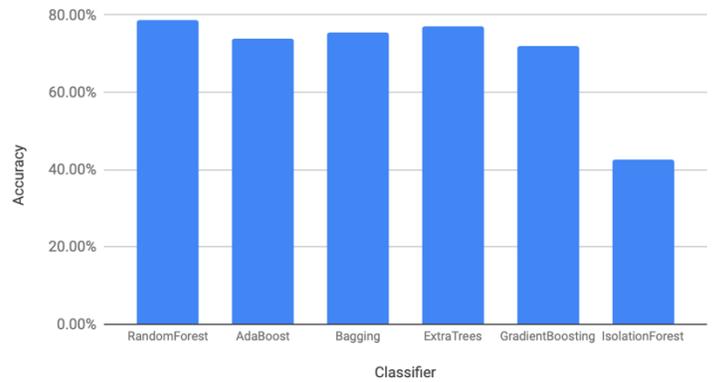

Figure 9. Bar Graph portraying the accuracy score for each ensemble classifier.

After lots of manipulation, for the Random Forest Classifier ML algorithm the parameters of n_estimators: [4, 6, 9, 13, 8], max_features : ['log2, 'sqrt', 'auto'], criterion: ['entropy',' gini'], max_depth: [1,16, 32, 32, 26], min_samples_split: [2, 3, 5,8,12], and min_samples_leaf: [1, 2, 8, 10, 15] were used.

This resulted in an 81.97% accuracy for the ML algorithm and though K-fold cross validation, a mean accuracy of 82.13%. Since the values are very similar it can be considered that the ML algorithm has an accuracy rate of roughly 82%. Furthermore, the DNN was trained with 350 epochs with a batch size of 8. The model was coded using 2 hidden layers with 8 and 5 nodes respectively. The activations "relu" and "sigmoid"' were used for the layers of the DNN. Adding on, the DNN resulted in a 92% accuracy as determined by the accuracy rate and the K-fold cross validation.

At first for the DNN, a low accuracy score was obtained that was below the criteria established earlier (75%). To better meet the performance criteria, this outcome required changes to the design and mechanics of DNN. Changes to the epochs were made and, as tested, up till a certain period, it resulted in much higher accuracy rates of the neural network. This can be portrayed below in Figure 10.

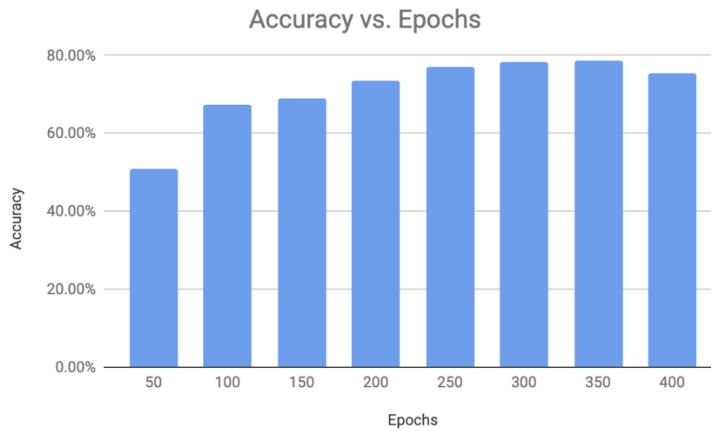

Figure 10. Bar Graph delineating relationship between accuracy and epochs.

## V. Conclusion and Future Steps

In general, it can be said the more iterations a network goes through the better it learns, but this idea has limitations. This concept, known as "early stopping," explains why as the epochs increased, the accuracy only increased up to a certain point and then after that any increases resulted in a decrease in accuracy rates. At that point in training, the network stopped generalizing and start learning the statistical noise in the training dataset [6]. Future steps include improving model performance by obtaining additional data and turning the current web application into an IOS app.